\documentclass[conference]{IEEEtran}
\IEEEoverridecommandlockouts
\usepackage{colortbl}
\usepackage{cite}
\usepackage{amsmath,amssymb,amsfonts}
\usepackage{algorithmic}
\usepackage{graphicx}
\usepackage{textcomp}
\usepackage{xcolor}
\usepackage{multirow}
\usepackage{svg}
\usepackage{subcaption}
\usepackage{hyperref}
\usepackage{url}
\usepackage{xspace}
\usepackage{xcolor}
\usepackage{makecell}
\usepackage{array}

\def\BibTeX{{\rm B\kern-.05em{\sc i\kern-.025em b}\kern-.08em
    T\kern-.1667em\lower.7ex\hbox{E}\kern-.125emX}}
\begin{document}

\newcommand{\method}{AttrSyn\xspace}
\newcommand\lx[1]{\textcolor{blue}{Linxin: #1}}
\newcommand\todo[1]{\textcolor{red}{ToDo: #1}}

\title{Attributed Synthetic Data Generation for Zero-shot Domain-specific Image Classification}

\author{Shijian Wang\textsuperscript{1}, Linxin Song\textsuperscript{2,3}, Ryotaro Shimizu\textsuperscript{3,4}, Masayuki Goto\textsuperscript{2}, Hanqian Wu\thanks{\textsuperscript{\dag}Corresponding author}\textsuperscript{\dag~1}\\
\small
\textsuperscript{1}Southeast University, China, \textsuperscript{2}Waseda University, Japan\\
\textsuperscript{3}ZOZO Research, Japan, \textsuperscript{4}University of California San Diego, USA\\
{\tt\small shijian@seu.edu.cn,
songlx.imse.gt@ruri.waseda.jp}\\
{\tt\small r2shimizu@ucsd.edu, masagoto@waseda.jp, hanqian@seu.edu.cn}
}

\maketitle

\begin{abstract}
Zero-shot domain-specific image classification is challenging in classifying real images without ground-truth in-domain training examples. Recent research involved knowledge from texts with a text-to-image model to generate in-domain training images in zero-shot scenarios. However, existing methods heavily rely on simple prompt strategies, limiting the diversity of synthetic training images, thus leading to inferior performance compared to real images. In this paper, we propose AttrSyn, which leverages large language models to generate attributed prompts. These prompts allow for the generation of more diverse attributed synthetic images. Experiments for zero-shot domain-specific image classification on two fine-grained datasets show that training with synthetic images generated by AttrSyn significantly outperforms CLIP's zero-shot classification under most situations and consistently surpasses simple prompt strategies.
\end{abstract}


\vspace{-1mm}
\section{Introduction}

Data scarcity poses a significant challenge in the field of domain-specific image classification~\cite{deng2009imagenet, mann2020language, radford2019language, schuhmann2022laion}, as the scarcity of high-quality in-domain data hinders the development of robust image classification systems.
In particular, zero-shot domain-specific image classification~\cite{wang2019survey, menon2022visual, pratt2023does}, which refers to classifying real images without having access to ground-truth in-domain training examples, emerges as a crucial technique to this dilemma.

Breakthroughs in text-to-image models, especially diffusion models~\cite{yang2023diffusion, nichol2021improved, song2020score, rombach2022high}, have enabled the efficient generation of a vast number of high-quality synthetic images. Recent works~\cite{cai2024uncertainty, li2024synthetic, he2022synthetic, fan2023scaling, gowda2023synthetic} are exploring the potential of employing synthetic images to tackle the challenges associated with zero-shot domain-specific image classification.
However, the majority of these efforts primarily focus on how synthetic images are utilized during the training stage while overlooking the intrinsic issues in synthetic images. Most relevant research\cite{peng2015learning, chen2019learning, ros2016synthia, abu2018augmented, yen2022nerf} leverages simple class-conditional prompts for text-to-image models to generate synthetic images. Such simple prompt strategies inherently limit the diversity of the synthetic images, which is critical to reducing the gap between synthetic and real images when used as in-domain training data~\cite{shipard2023diversity, fan2023scaling, yu2023diversify, yang2017diversity}.

Inspired by ~\cite{yu2024large}, which introduces an attribute-based text generation approach to enhance text classification by enriching the diversity of the textural training data, we propose \textbf{\method}, designed to generate high-quality in-domain synthetic training images according to a diverse attribute library.
Specifically, we obtain the attribute concepts and their corresponding attribute candidate values for a given dataset in an interactive, semi-automatic manner with the assistance of large language models (LLMs). 
We randomly combine these attributes with the class name to create attributed image generation prompts. These attributed prompts are then fed into a text-to-image model, e.g., Stable Diffusion~\cite{rombach2022high}, to generate attributed synthetic images.

To demonstrate the effectiveness of \method, we train classifiers with synthetic images produced by \method and simple prompt strategies and compare their performances on two domain-specific image classification datasets CUB-200~\cite{birds2011} and CUB-200-painting~\cite{wang2020progressive}, which are particularly challenging for zero-shot classification.
Additionally, we include CLIP~\cite{radford2021learning} zero-shot classification as a strong baseline to highlight the improvements achieved by \method in generating diverse and domain-specific training images.
The experimental results show that training on the dataset generated by \method can consistently outperform simple prompt strategies across all settings and significantly surpasses CLIP's zero-shot performance under most situations on both fine-grained datasets, yielding performance enhancements ranging up to a maximum of \textbf{13.62\%}.
We summarize our contributions as follows:
\begin{itemize}
    \item We propose \textbf{\method}, a novel attributed synthetic image generation method to facilitate zero-shot domain-specific image classification tasks.
    % \item We highlight the significance of focusing on the upstream generation process of synthetic images.
    \item We demonstrate the superiority of \textbf{\method} over simple prompt strategies and CLIP zero-shot classification through extensive experiments, highlighting its potential for improving zero-shot domain-specific image classification's performance.
\end{itemize}

\begin{figure*}[!t]
  \centering
  \includegraphics[width=\linewidth]{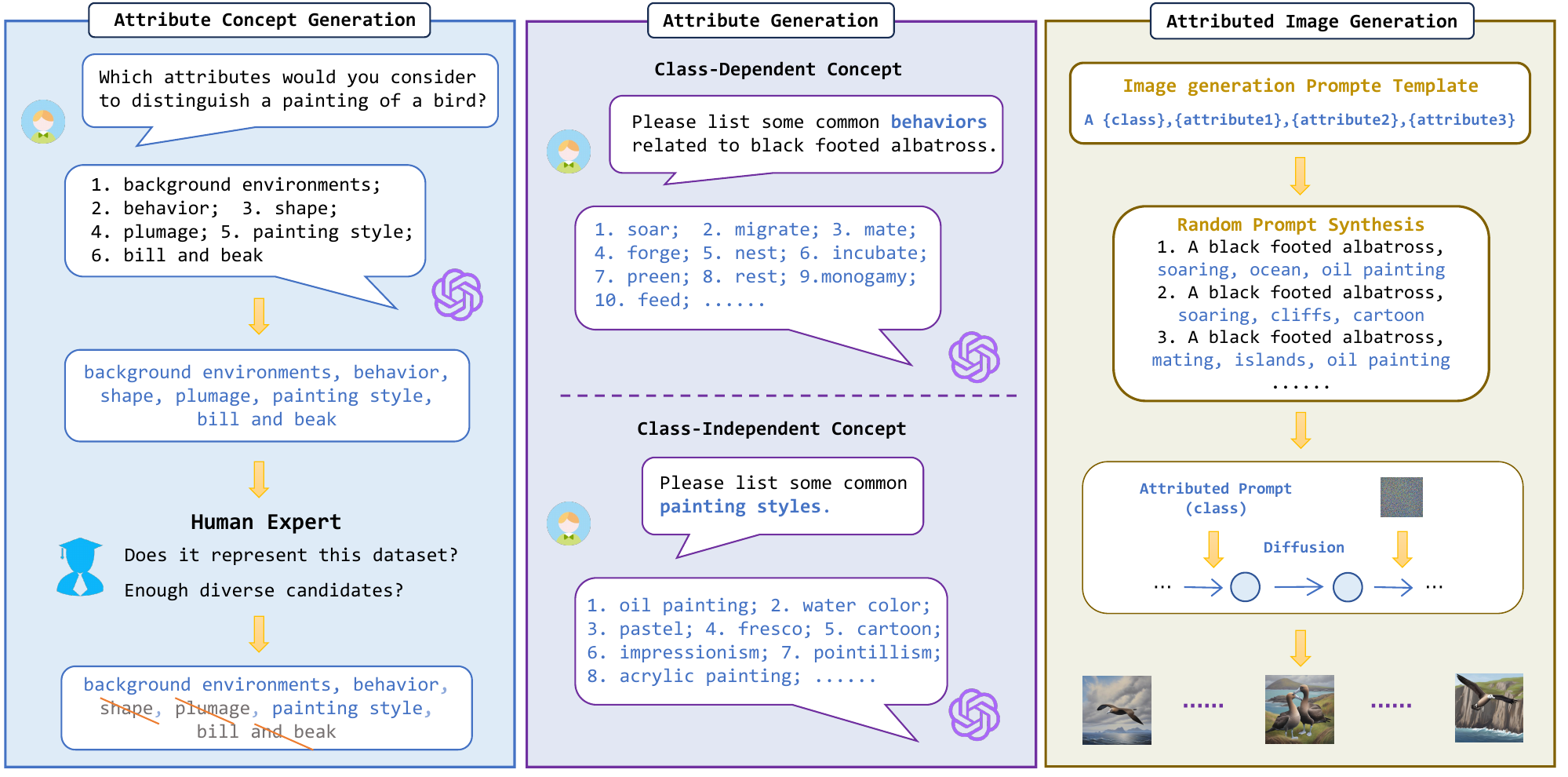}
  \caption{Overall workflow of \method. In the attribute concept generation stage, for a given dataset, high-quality attribute concepts are derived by querying a large language model and human interactive filtering. In the attribute generation stage, the obtained attribute concepts are categorized into class-dependent and class-independent concepts, and different query strategies are adopted to generate diverse attribute candidate values. In the attributed image generation stage, attribute candidate values from various attribute concepts are randomly selected and combined with class names to create diverse attributed prompts, which are subsequently sent to a text-to-image model to produce the corresponding attributed images.}
  \label{fig:workflow}
\end{figure*}

\vspace{-2mm}
\section{Related Work}
\label{sec:related}

\noindent \textbf{Zero-shot domain-specific image classification.}
In traditional zero-shot learning (ZSL) for domain-specific image classification, a strict distinction is maintained between seen and unseen classes~\cite{wang2019survey}. This line of work typically relies on learning intermediate attribute classifiers~\cite{rohrbach2011evaluating,lampert2013attribute} and modeling seen class proportions~\cite{zhang2015zero}. However, the introduction of CLIP~\cite{radford2021learning} has extended the zero-shot paradigm by eliminating the need for in-domain training data. CLIP jointly trains image and text encoders to learn shared representations, enabling zero-shot predictions by matching image embeddings with the most similar descriptive text embeddings. Consistent with CLIP, our zero-shot setting also eliminates the need for any in-domain training data, diverging from traditional ZSL methodologies.

\noindent \textbf{Learn from synthetic images.}
Synthetic training data has been proven effective in enhancing performance across various domains~\cite{bauer2024comprehensive}. With advancements in text-to-image generation models~\cite{yang2023diffusion, ho2020denoising, nichol2021improved, song2020score, rombach2022high}, particularly diffusion models like the DALL-E series~\cite{ramesh2021zero}, synthetic images have become widely utilized in numerous computer vision tasks~\cite{cai2024uncertainty, he2022synthetic, fan2023scaling, gowda2023synthetic}. Furthermore, significant efforts have been made to develop methods for learning transferable representations from synthetic images~\cite{li2024synthetic}. However, these approaches are mostly constrained by the limited diversity of synthetic data, which creates a significant gap compared to real data. To address this limitation, our work introduces an attribute-based method aimed at increasing the diversity of synthetic images, thereby improving their effectiveness for model training.

\vspace{-1mm}
\section{Generating Attributed Synthetic Images}
\label{sec:method}

In this section, we present the details of \method, which leverages attribute-based synthetic data generation to achieve zero-shot domain-specific image classification. Specifically, \method employs large language models to produce various plausible attributes that enhance the image generation prompts fed to the text-to-image model. These enhanced prompts enable more effective synthetic in-domain training data for zero-shot domain-specific image classification. The overall workflow is shown in Fig.~\ref{fig:workflow}.

\vspace{-1mm}
\subsection{Datasets}
Given the challenges and limited exploration of domain-specific fine-grained image classification in a zero-shot setting, we utilize two fine-grained image classification datasets spanning the \textit{photo} and \textit{painting} domains.

\begin{itemize}
    \item \textbf{CUB-200}~\cite{birds2011}: The CUB-200 dataset contains 200 different categories of bird photos, capturing variations in appearance, pose, and background.
    \item \textbf{CUB-200-Painting}~\cite{wang2020progressive}: The CUB-200-Painting dataset contains 200 different categories of bird paintings, and these categories are consistent with CUB-200.
\end{itemize}

\noindent We summarize the statistics of the two fine-grained image classification datasets in Tab.~\ref{tab:datasets}.
Without using any real training images, we generate synthetic images to train classifiers and evaluate on the corresponding real image test sets.
% from which we can see that the training set size of CUB-200 is nearly 6,000 and CUB-200-Painting doesn't split the training set. Therefore, for both datasets, we generate \textbf{6,000} synthetic images to train classifiers and test on their test sets.

\vspace{-1mm}
\subsection{Attribute Concept Generation}

To obtain diverse and high-quality synthetic training data for zero-shot image classification, we adopt a semi-automated human-machine collaborative attribute generation approach inspired by ~\cite{yu2024large}.
Our method initiates by defining several attribute concepts for a given dataset. 
Since the process demands a comprehensive understanding of the dataset, particularly in zero-shot scenarios where real data is unavailable, we employ a large language model $\mathcal{L}$ that has a wealth of world knowledge to assist in initially determining attribute concepts $\mathcal{C} = \{c_i \mid i = 1, \ldots, m\}$, which consists of $m$ raw attribute concepts.
Specifically, we selected \emph{GPT-4}\footnote{\url{https://platform.openai.com/docs/models/gp}}~\cite{achiam2023gpt} as the large language model for querying, given its strong instruction-following capabilities and extensive knowledge. 
To identify potential attribute concepts, we posed queries such as: \textit{“Which attributes would you consider to distinguish a painting of a bird?”} for the CUB-200-Painting dataset, resulting in responses like \textit{“background environments, behavior, shape, plumage, painting style, bill and beak”}.

Human experts then interactively selects $n$ high-quality attribute concepts $\mathcal{C}' = \{c'_i \in \mathcal{C} \mid i = 1, \ldots, n, \forall r \in \mathcal{R}, \phi_r(c'_i) = 1\}$,
where $n \leq m$, $\phi_r(\cdot)\in \{0, 1\}$ is a boolean function that evaluates whether an attribute concept $c'_i$ satisfies rule $r$, and $\mathcal{R}$ is a set of rules considered by human experts.
Specifically, human experts employed two rules for filtering attribute concepts: 
(i) ensuring that attribute concepts captured distinguishable features within the dataset to maintain \textbf{quality}, and 
(ii) verifying sufficient variation in candidate values to ensure \textbf{diversity}. 
As illustrated in Fig.~\ref{fig:workflow}, in the case of the CUB-200-Painting dataset, the \textit{shape} attribute concept was excluded due to its low distinguishability, whereas the \textit{plumage} and \textit{bill and beak} attribute concepts were omitted because they lacked sufficient variation in their candidate values.

\begin{table}[t]
\caption{Statistics of datasets.}
\centering
\renewcommand{\arraystretch}{1.25}
\resizebox{\columnwidth}{!}{%
\begin{tabular}{cccccc}
\hline
\rowcolor{lightgray}
\textbf{Dataset} & \textbf{Domain} & \textbf{Task} & \textbf{\# Train} & \textbf{\# Test} & \textbf{\# Class} \\ \hline
CUB-200     & Photo           & Multi-class           & 5,994   & 5,794             & 200            \\
CUB-200-Painting & Painting        & Multi-class           &—   & 3,047             & 200            \\ \hline
\end{tabular}%
}
\label{tab:datasets}
\end{table}

\begin{table}[t]
\caption{Attribute concepts of datasets. \# Configs denotes the number of diversity configurations. The examples of attribute values can be found in Appendix A.}
\centering
\renewcommand{\arraystretch}{1.4}
\resizebox{\columnwidth}{!}{%
\begin{tabular}{cccc}
\hline
\rowcolor{lightgray}
\textbf{Dataset}                  & \textbf{\# Configs} & \textbf{Attr. Concept}     & \textbf{\# Attr. Values} \\ \hline
\multirow{3}{*}{CUB-200}     & \multirow{3}{*}{25,000}               & behavior               & 1,000                                   \\
                                  &                                    & background environment & 1,000                                   \\
                                  &                                    & photo style            & 5                                   \\ \hline
\multirow{3}{*}{CUB-200-Painting} & \multirow{3}{*}{25,000}               & behavior               & 1,000                                   \\
                                  &                                    & background environment & 1,000                                   \\
                                  &                                    & painting style         & 5                                   \\ \hline
\end{tabular}%
}
\label{tab:attributes}
\end{table}

\vspace{-1mm}
\subsection{Attribute Generation}

To query diverse fine-grained attributes according to each attribute concept, we further query the \textit{GPT-4} to seek diverse attribute candidate values.
We categorize attribute concepts $\mathcal{C'}$ into class-dependent and class-independent concepts and apply different generation strategies for each.
Class-dependent concepts exhibit diverse attribute values among different classes, such as the \emph{``background environment"} attribute concept for CUB-200-Painting.
Class-independent concepts, such as \emph{``painting style"} for CUB-200-Painting, can share the same values across all classes.
For each concept $c'_i \in \mathcal{C'}$, we leverage the \textit{GPT-4} to generate a set of attribute values $\mathcal{V}_i = \{v_{i,j} \mid j = 1, \ldots, k_i\}$, where $k_i$ denotes the number of attribute values for concept $c'_i$. We can quantitatively measure the overall diversity by $\prod_{i=1}^{n} k_i$, the number of all possible attribute combinations, where $n$ is the number of all attribute concepts.
The following are the specific strategies for generating attribute values for class-dependent and class-independent concepts.

\noindent \textbf{Class-dependent concepts.}
For class-dependent concepts, we utilize prompts incorporating the class name to query large language models. 
The query prompt template we use is 
``\emph{Please list some common \{attribute concept\} related to the \{class name\}.}"
where the placeholders \{\emph{attribute concept}\} and \{\emph{class name}\} are replaced with the specific attribute concept and class, respectively. 

\noindent \textbf{Class-independent concepts.}
We use a generic prompt to acquire class-independent attribute values applicable across various classes. 
The query prompt template is 
``\emph{Please list some common \{attribute concept\}.}"
where the placeholders \{\emph{attribute concept}\} are replaced with the specific concept. 

As presented in Tab.~\ref{tab:attributes}, we generated and selected 3 attribute concepts for each dataset, with each attribute concept containing 5 attribute values for each class. This resulted in a total of 125 unique diversity configurations per class. Since each dataset consists of 200 classes, this yielded a total of 25,000 diversity configurations for each dataset. The examples of these attribute values can be found in Appendix A.
In this study, we focus on leveraging attributes to enhance the diversity of synthetic training data. The search for optimal diversity configurations is left to future research.

\begin{figure}[t]
  \centering

  \begin{subfigure}{\linewidth}
    \centering
    \includegraphics[width=0.16\linewidth]{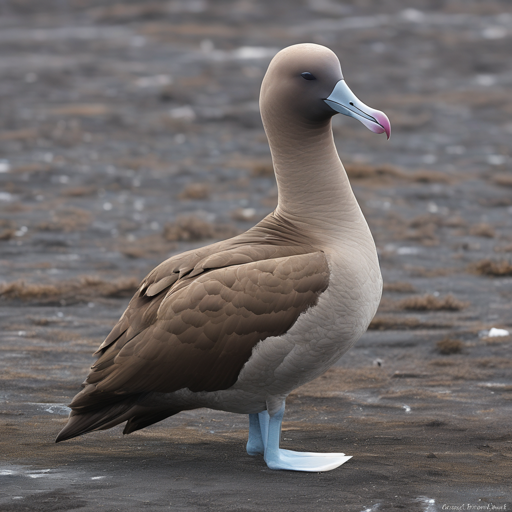}
    \hspace{-5pt}
    \includegraphics[width=0.16\linewidth]{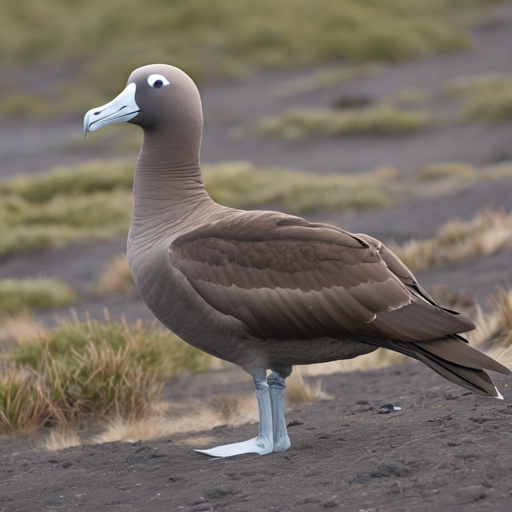}
    \hspace{-5pt}
    \includegraphics[width=0.16\linewidth]{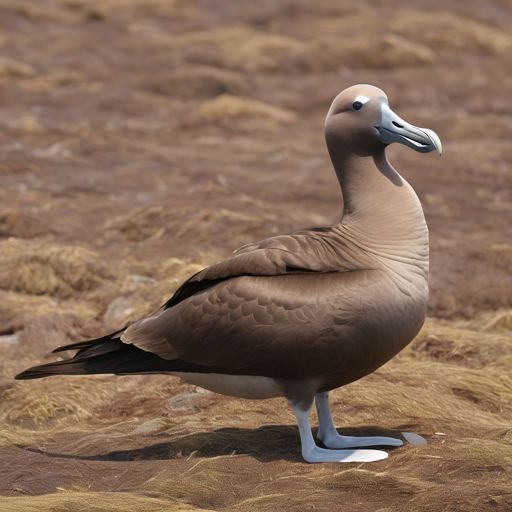}
    \hspace{-5pt}
    \includegraphics[width=0.16\linewidth]{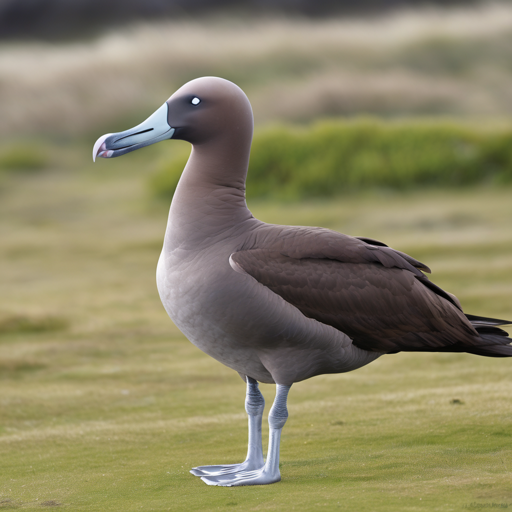}
    \hspace{-5pt}
    \includegraphics[width=0.16\linewidth]{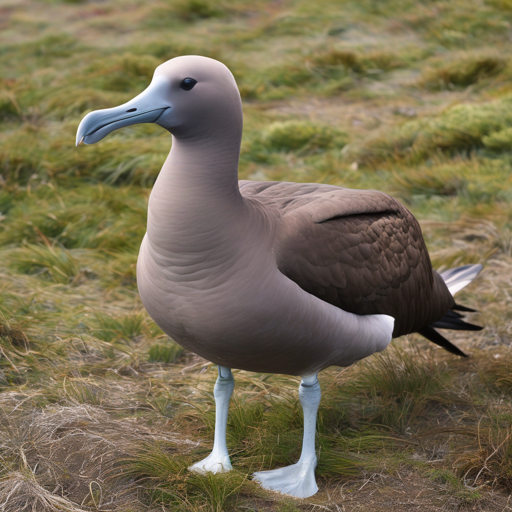}
    \hspace{-5pt}
    \includegraphics[width=0.16\linewidth]{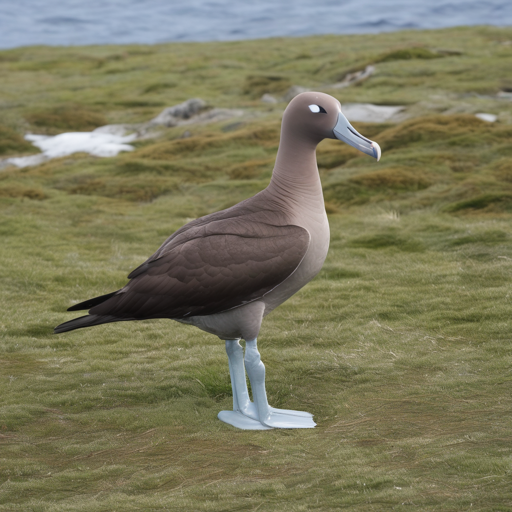}
    \hspace{-5pt}
    \includegraphics[width=0.16\linewidth]{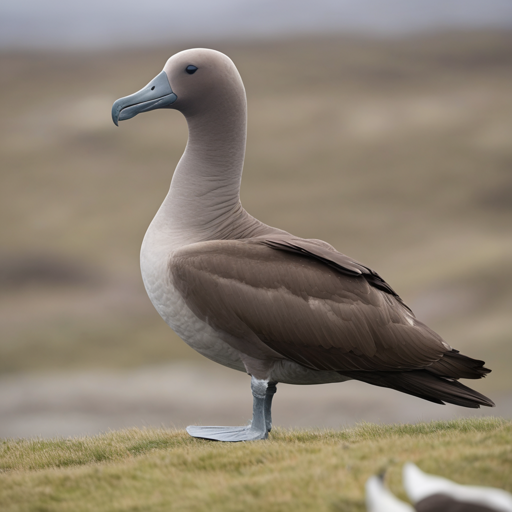}
    \hspace{-5pt}
    \includegraphics[width=0.16\linewidth]{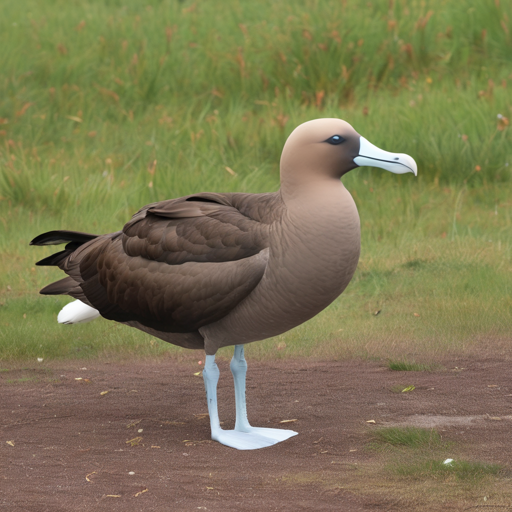}
    \hspace{-5pt}
    \includegraphics[width=0.16\linewidth]{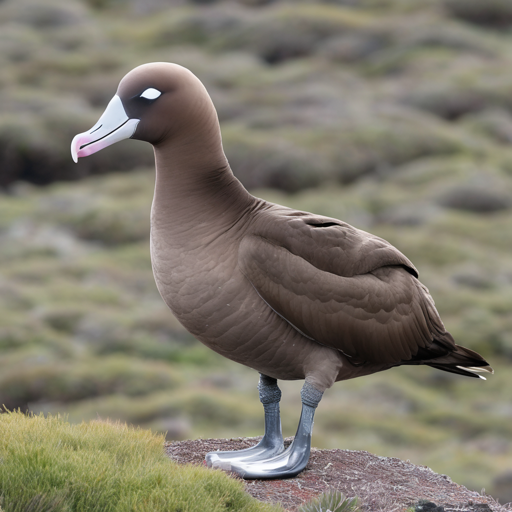}
    \hspace{-5pt}
    \includegraphics[width=0.16\linewidth]{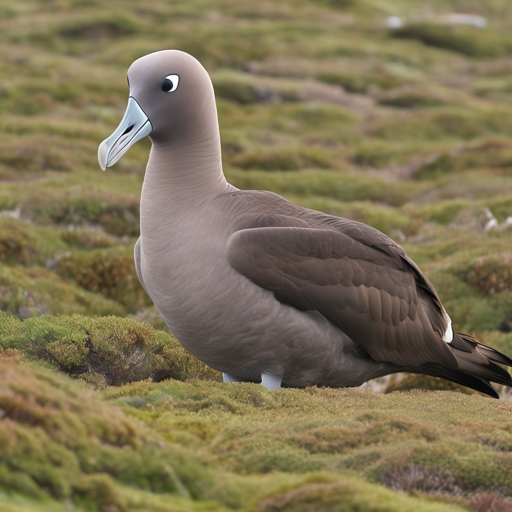}
    \hspace{-5pt}
    \includegraphics[width=0.16\linewidth]{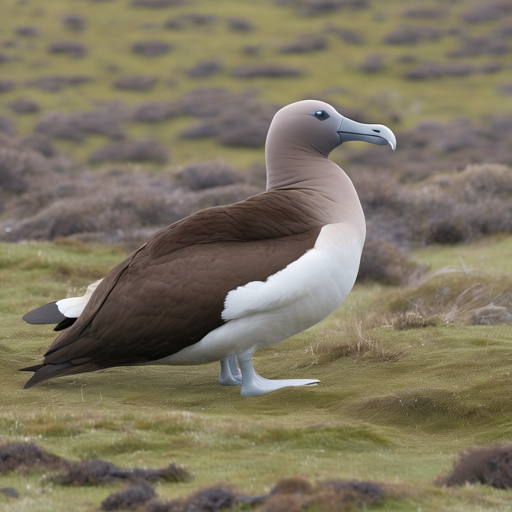}
    \hspace{-5pt}
    \includegraphics[width=0.16\linewidth]{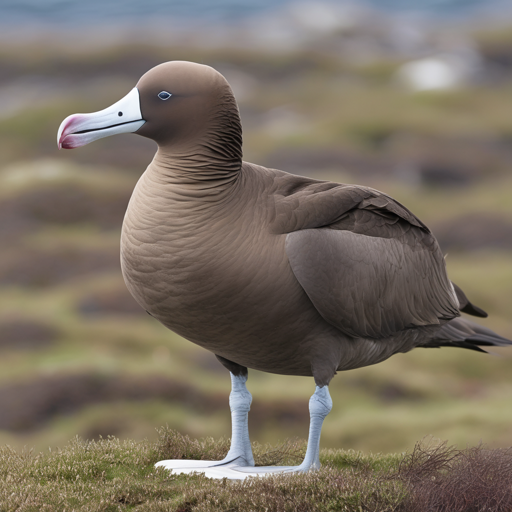}
    \caption{\textit{photos} generated by base prompt.}
    \vspace{3pt}
  \end{subfigure}

  \begin{subfigure}{\linewidth}
    \centering
    \includegraphics[width=0.16\linewidth]{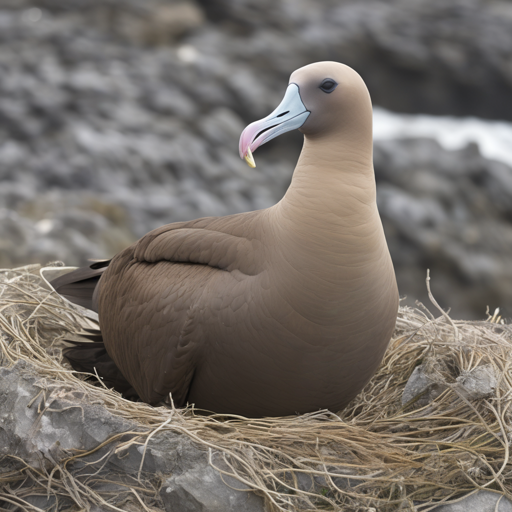}
    \hspace{-5pt}
    \includegraphics[width=0.16\linewidth]{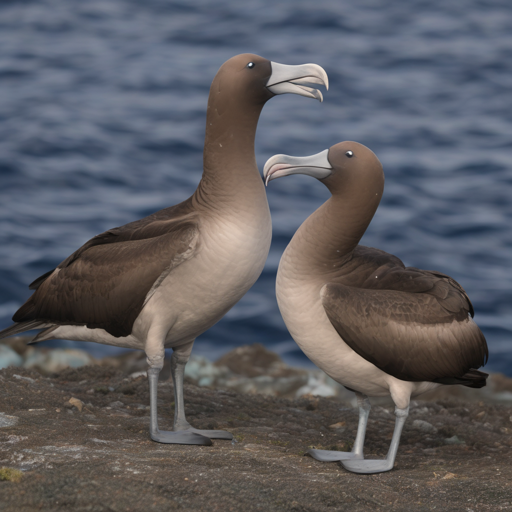}
    \hspace{-5pt}
    \includegraphics[width=0.16\linewidth]{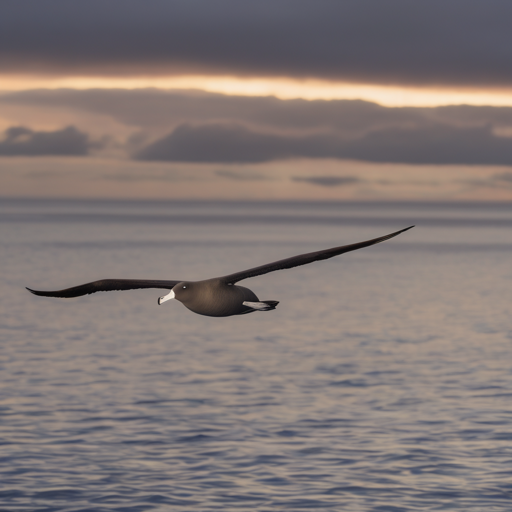}
    \hspace{-5pt}
    \includegraphics[width=0.16\linewidth]{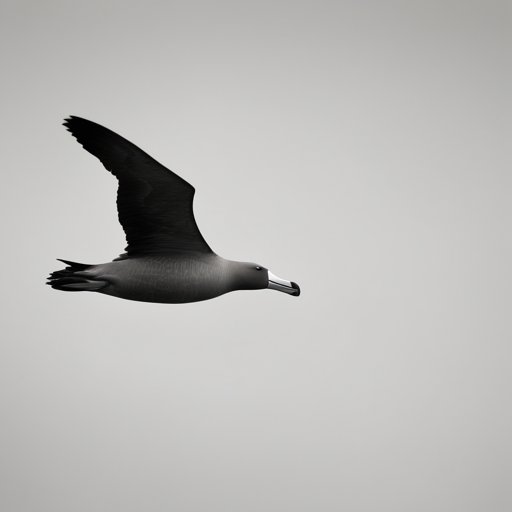}
    \hspace{-5pt}
    \includegraphics[width=0.16\linewidth]{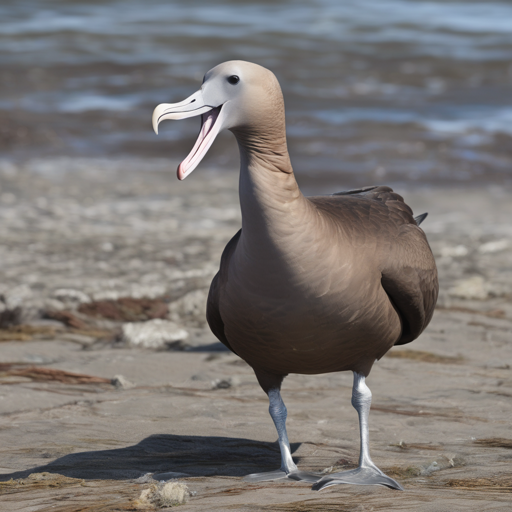}
    \hspace{-5pt}
    \includegraphics[width=0.16\linewidth]{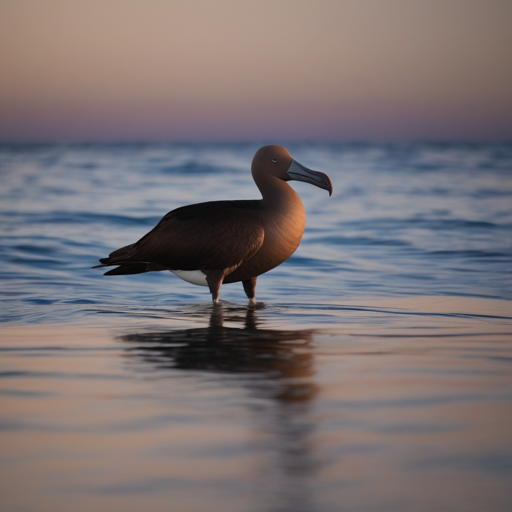}
    \hspace{-5pt}
    \includegraphics[width=0.16\linewidth]{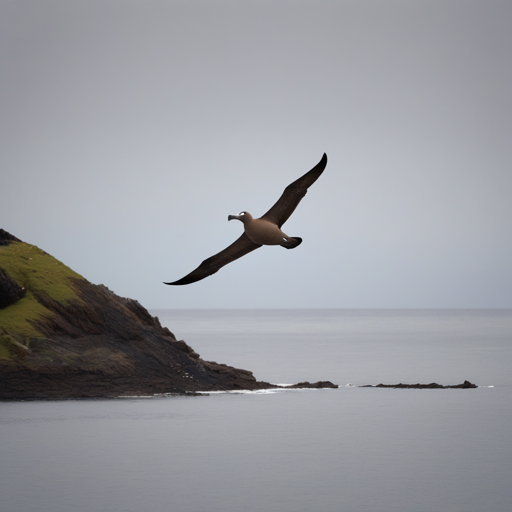}
    \hspace{-5pt}
    \includegraphics[width=0.16\linewidth]{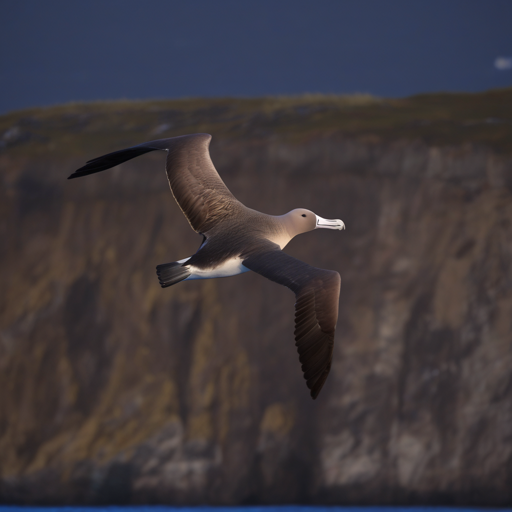}
    \hspace{-5pt}
    \includegraphics[width=0.16\linewidth]{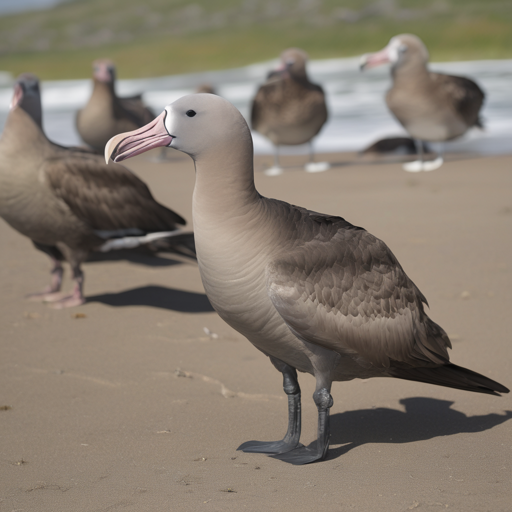}
    \hspace{-5pt}
    \includegraphics[width=0.16\linewidth]{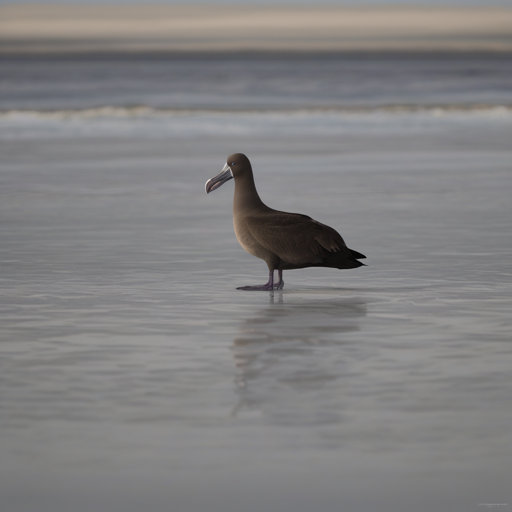}
    \hspace{-5pt}
    \includegraphics[width=0.16\linewidth]{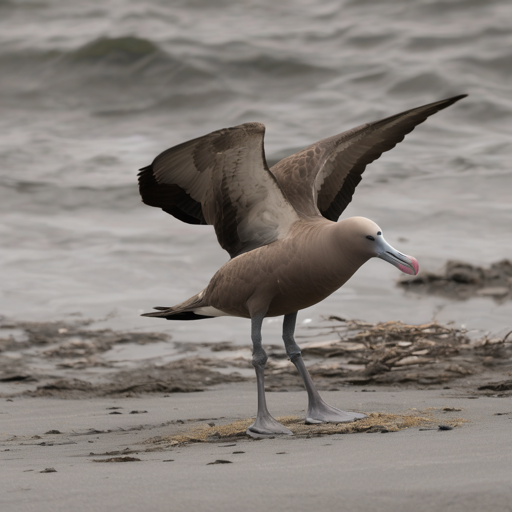}
    \hspace{-5pt}
    \includegraphics[width=0.16\linewidth]{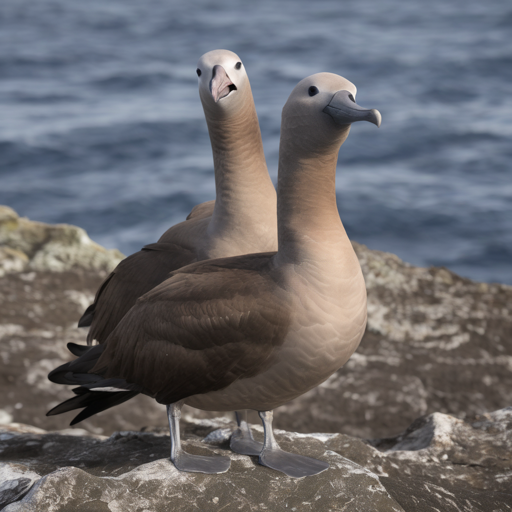}
    \caption{\textit{photos} generated by \textbf{\method} prompt.}
    \vspace{3pt}
  \end{subfigure}

  \begin{subfigure}{\linewidth}
    \centering
    \includegraphics[width=0.16\linewidth]{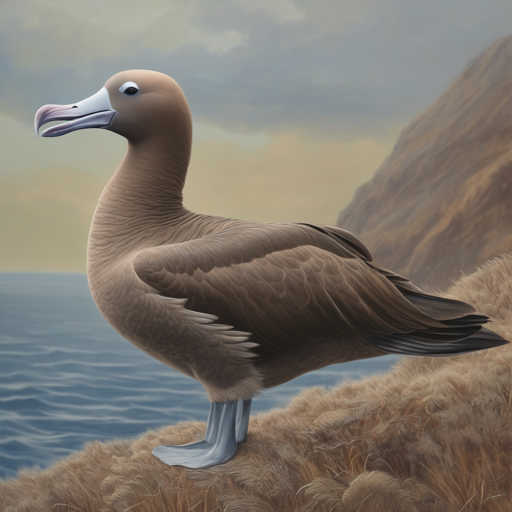}
    \hspace{-5pt}
    \includegraphics[width=0.16\linewidth]{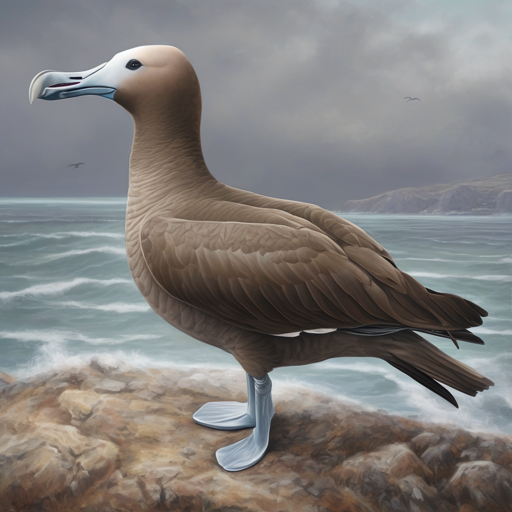}
    \hspace{-5pt}
    \includegraphics[width=0.16\linewidth]{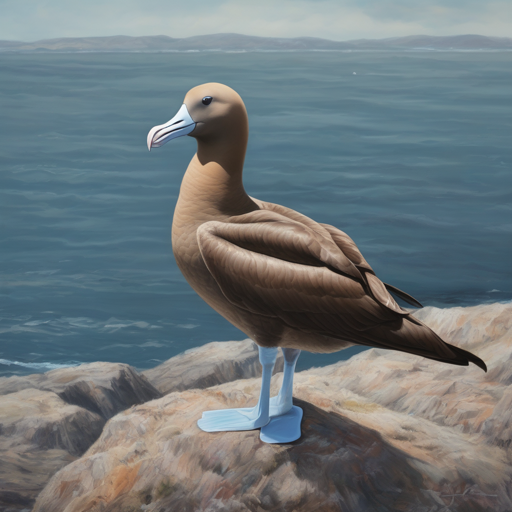}
    \hspace{-5pt}
    \includegraphics[width=0.16\linewidth]{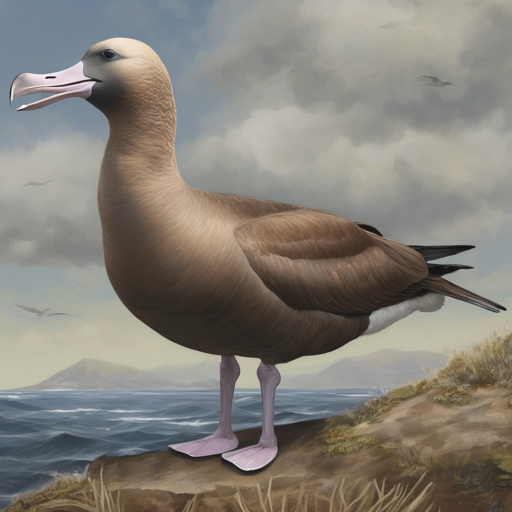}
    \hspace{-5pt}
    \includegraphics[width=0.16\linewidth]{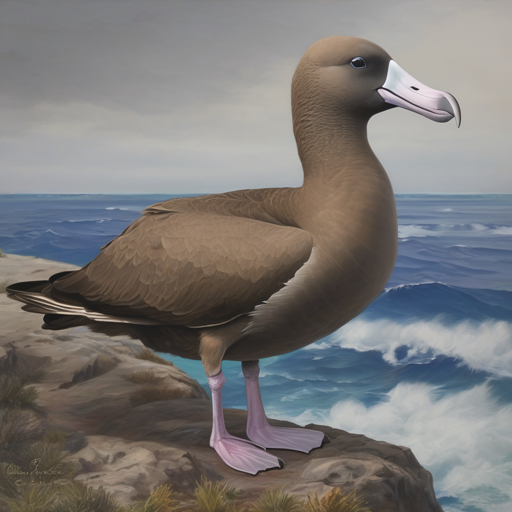}
    \hspace{-5pt}
    \includegraphics[width=0.16\linewidth]{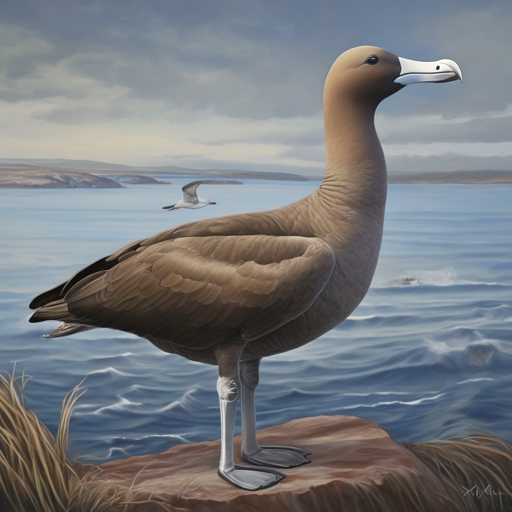}
    \hspace{-5pt}
    \includegraphics[width=0.16\linewidth]{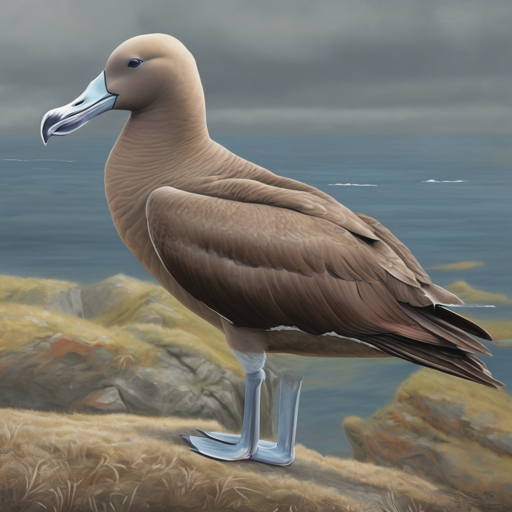}
    \hspace{-5pt}
    \includegraphics[width=0.16\linewidth]{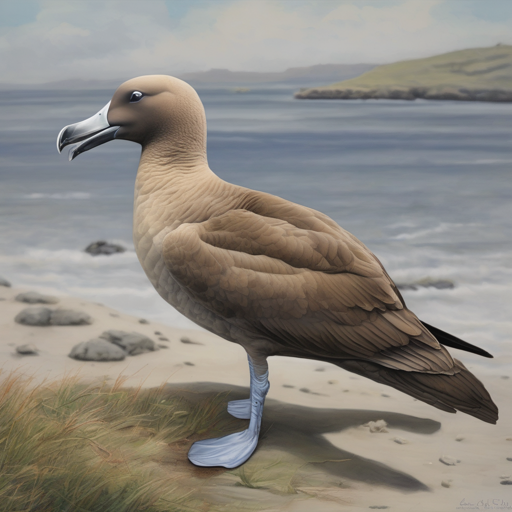}
    \hspace{-5pt}
    \includegraphics[width=0.16\linewidth]{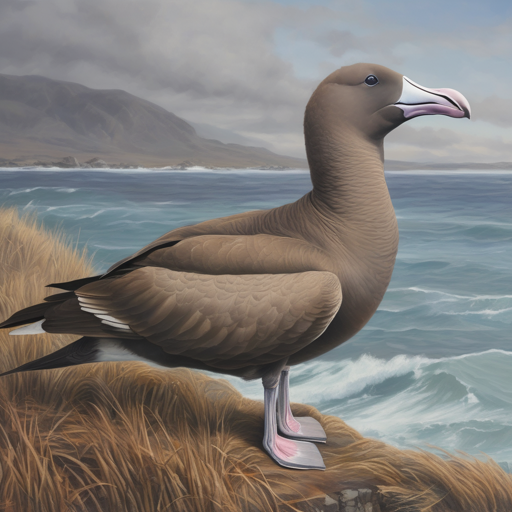}
    \hspace{-5pt}
    \includegraphics[width=0.16\linewidth]{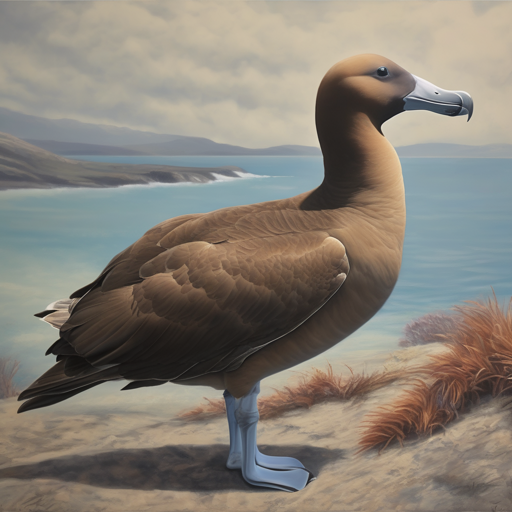}
    \hspace{-5pt}
    \includegraphics[width=0.16\linewidth]{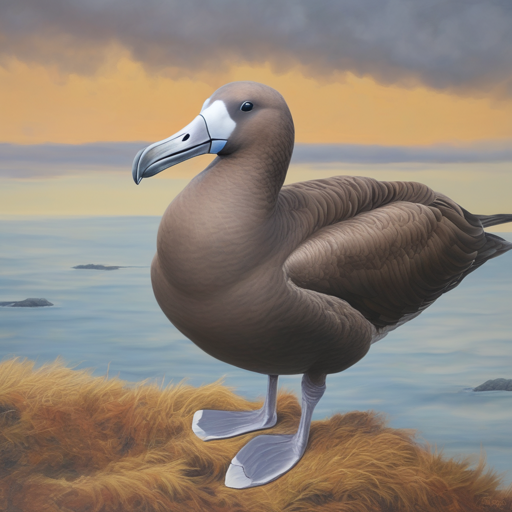}
    \hspace{-5pt}
    \includegraphics[width=0.16\linewidth]{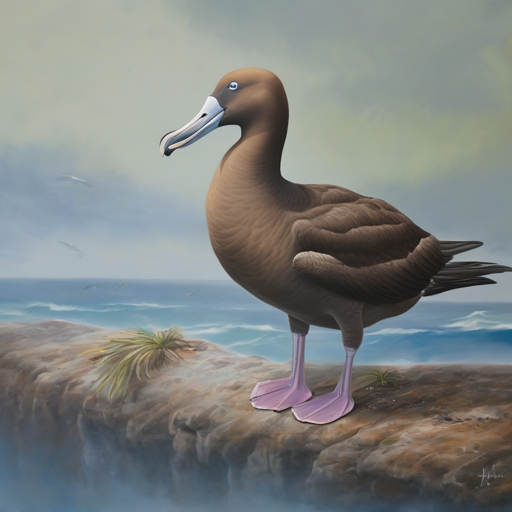}
    \caption{\textit{paintings} generated by base prompt.}
    \vspace{3pt}
  \end{subfigure}

  \begin{subfigure}{\linewidth}
    \centering
    \includegraphics[width=0.16\linewidth]{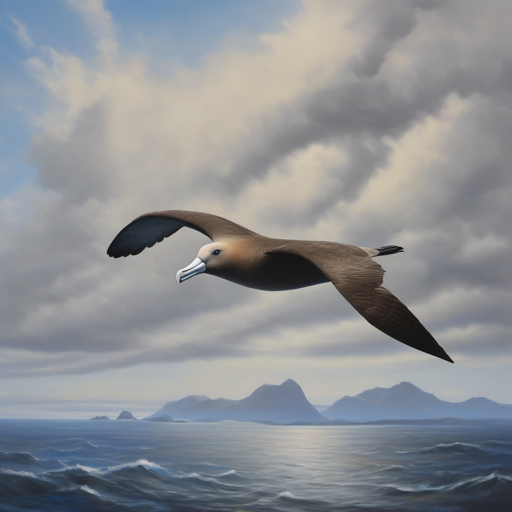}
    \hspace{-5pt}
    \includegraphics[width=0.16\linewidth]{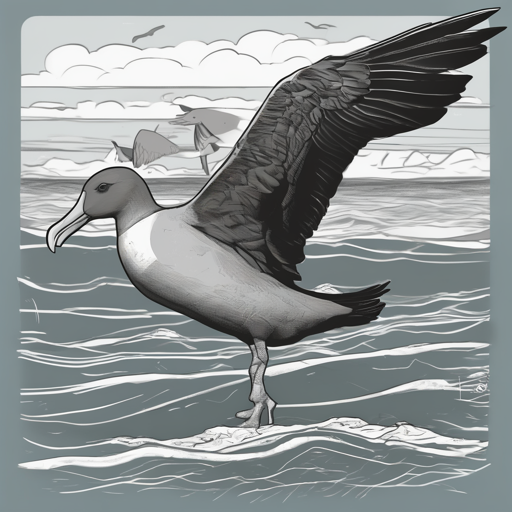}
    \hspace{-5pt}
    \includegraphics[width=0.16\linewidth]{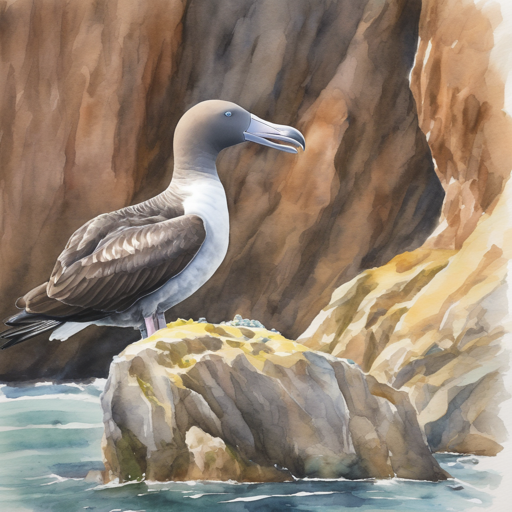}
    \hspace{-5pt}
    \includegraphics[width=0.16\linewidth]{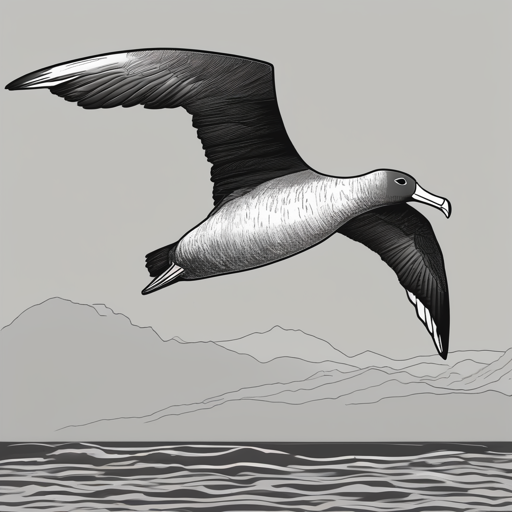}
    \hspace{-5pt}
    \includegraphics[width=0.16\linewidth]{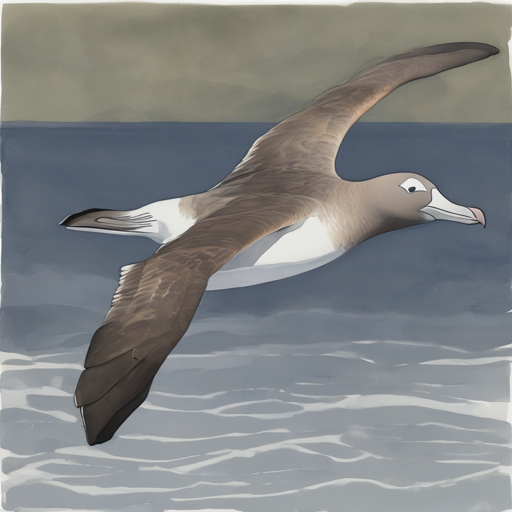}
    \hspace{-5pt}
    \includegraphics[width=0.16\linewidth]{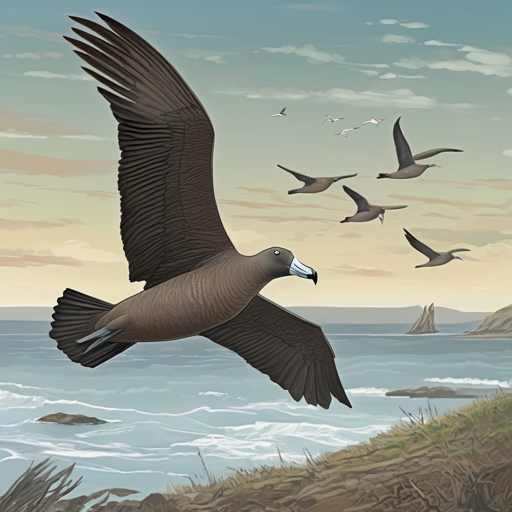}
    \hspace{-5pt}
    \includegraphics[width=0.16\linewidth]{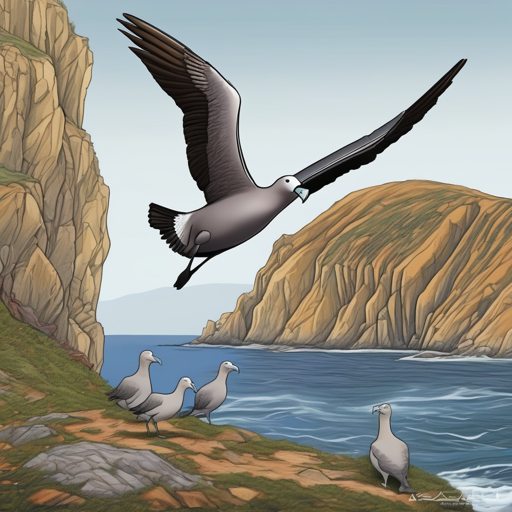}
    \hspace{-5pt}
    \includegraphics[width=0.16\linewidth]{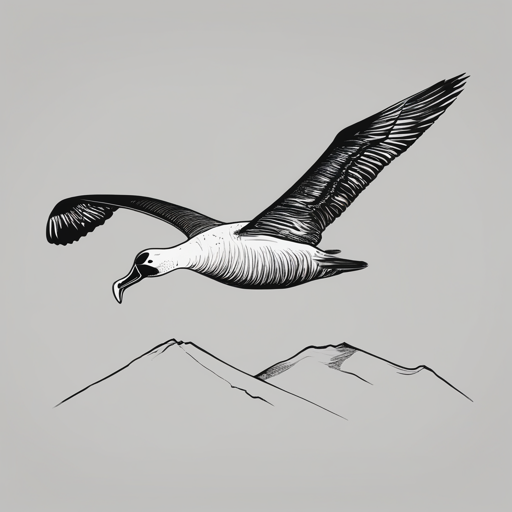}
    \hspace{-5pt}
    \includegraphics[width=0.16\linewidth]{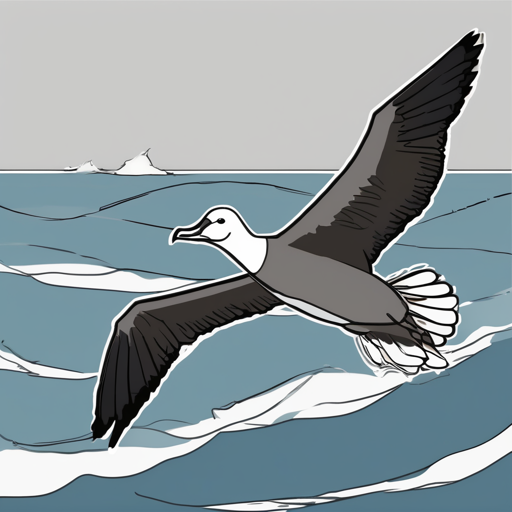}
    \hspace{-5pt}
    \includegraphics[width=0.16\linewidth]{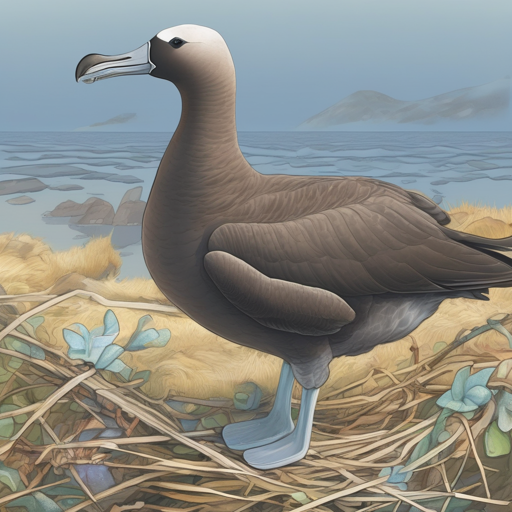}
    \hspace{-5pt}
    \includegraphics[width=0.16\linewidth]{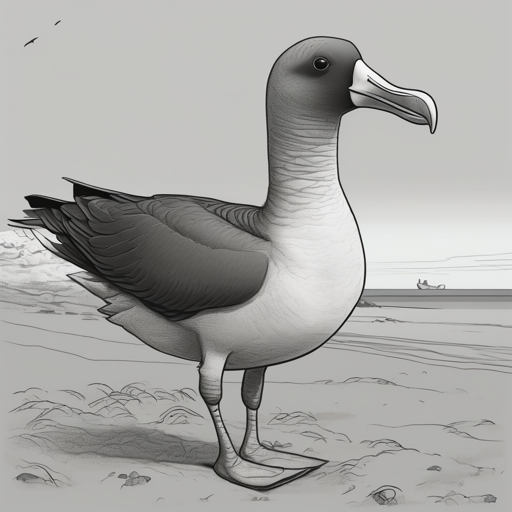}
    \hspace{-5pt}
    \includegraphics[width=0.16\linewidth]{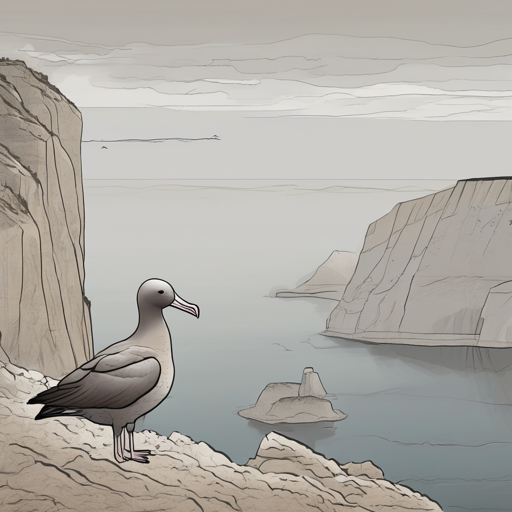}
    \caption{\textit{paintings} generated by \textbf{\method} prompt.}
  \end{subfigure}

  \caption{Visualization of synthetic images generated by the base prompt and \method for the black-footed albatross class. \method produces more diverse images compared to the base prompt in both the photo and painting domains.}
  \label{fig:visualization}
\end{figure}

\vspace{-1mm}
\subsection{Attributed Image Generation}
\label{sec:image-gen}

\noindent \textbf{Attributed prompt generation.}
To effectively prompt text-to-image models to produce images that display diversity across multiple attribute concepts, we randomly select one attribute value from each attribute concept and concatenate them with the class name to formulate attributed image generation prompts $ \mathcal{P} = \{y \oplus v_{1,j_1} \oplus v_{2,j_2} \oplus \cdots \oplus v_{n,j_n} \mid v_{i,j_i} \in \mathcal{V}_i, i = 1, \ldots, n\}$, where $\oplus$ represents the concatenation operation, $y$ is the class name, and $v_{i,j_i}$ is a randomly selected attribute value from the generated attribute values $\mathcal{V}_i$ for each attribute concept $c'_i \in \mathcal{C}'$.
For example, in the case of the \textit{black-footed albatross} class from the CUB-200-Painting dataset, one of the diversity configurations is \{\textit{behavior = soaring, background environment = ocean, painting style = oil painting}\}. This configuration can then be transformed into a complete prompt: “\textit{A black-footed albatross, soaring, ocean, oil painting}”.

\noindent \textbf{Image generation.}
These attributed prompts are subsequently sent into a text-to-image model $\mathcal{G}$, enabling the generation of diverse images $\mathcal{I}=\mathcal{G}(\mathcal{P})$ with specific attributes.
Specifically, considering the quality and controllability of synthetic images, we choose \emph{stable-diffusion-xl-base-1.0}\footnote{\url{https://huggingface.co/stabilityai/stable-diffusion-xl-base-1.0}}~\cite{podell2023sdxl}, a cutting-edge open-source diffusion-based generative model, as the experimental text-to-image model.
The primary factors influencing the text-to-image generation process are the \textit{image generation prompt}, \textit{guidance scale}, and \textit{generation timesteps}~\cite{shipard2023diversity, fan2023scaling}. Details of these factors can be found in Appendix B.
To isolate the impact of the image generation prompt, we fix the guidance scale at 5.0 and the generation timesteps at 50, thereby removing the influence of confounders.

\vspace{-1.5mm}
\subsection{Visualization}
\label{sec:vis}
To demonstrate the diversity of synthetic data generated by \method, we compare synthetic data generated by \method with a simple base prompt strategy. Specifically, we introduce the base prompt strategy and then visualize the data generated by the two methods for qualitative comparison.

\noindent \textbf{Base prompt.}
~\cite{shipard2023diversity} introduced a bag of tricks to enhance the diversity of synthetic images. We adopt the core tricks of their method as our baseline prompt strategy, which includes the domain and class name in the prompt, called the base prompt in our study. Specifically, for CUB-200, the base prompt is \emph{``a \{class name\} bird, photo"}, and we use \emph{``a \{class name\} bird, painting"} as the base prompt for the CUB-200-Painting dataset. For a fair comparison, the generation configuration used for generating synthetic images with the base prompt is consistent with AttrSyn, as detailed in Sec.~\ref{sec:image-gen}.

\noindent \textbf{A glimpse of the generated data.}
To qualitatively compare \method and the base prompt, we sample some synthetic images of \emph{black-footed albatross} generated by the two methods for visualization.
The visualization results are shown in Fig.~\ref{fig:visualization}, from which we can see that \method shows higher diversity than the base prompt across attributes such as background environment, behavior, and style for both datasets.

\vspace{-1.5mm}
\section{Training with Synthetic Images}
To quantitatively assess the effectiveness of \method in zero-shot domain-specific image classification, we train models exclusively on synthetic data and evaluate their performance on real image test sets. To isolate the impact of the synthetic data, we employ \textbf{linear probing}~\cite{he2022masked}, a well-established method for assessing the quality of training data and features, which avoids the confounding effects of model fine-tuning and optimization techniques. 
Specifically, we utilize CLIP backbone models to extract visual features from the synthetic images, train classifiers on these features, and evaluate their performance on the real image test sets.

\vspace{-1.5mm}
\subsection{Baseline}

We use the base prompt as a direct baseline and include CLIP zero-shot classification as a strong baseline. 
As discussed in Sec.~\ref{sec:related}, our zero-shot setting differs from traditional zero-shot learning, which typically relies on auxiliary attributes associated with predefined seen classes. Notably, our setting does not distinguish between seen and unseen classes and actually does not utilize any real image training data.
Therefore, we do not compare with traditional zero-shot methods.

\noindent \textbf{Synthetic images via the base prompt.} 
As shown in Sec.~\ref{sec:vis}, the synthetic data generated by \method shows higher diversity than the base prompt. Here we train on the synthetic data generated by the two methods to evaluate the impact of training data diversity on performance.

\noindent \textbf{CLIP zero-shot.} 
CLIP uses separate vision and language encoders to embed images and text into a shared space. It performs zero-shot classification by calculating the cosine similarity between an image embedding and predefined text prompt embeddings, assigning the class label with the highest similarity.
The CLIP text template we use is 
``\emph{a \{domain name\} of a \{class name\}}".

\vspace{-1.5mm}
\subsection{Experiment Setup}

\noindent \textbf{Synthetic training data.} 
As shown in Tab.~\ref{tab:datasets}, the real image training set size of CUB-200 is nearly 6,000 and CUB-200-Painting doesn't split the training set. 
To ensure consistency, for both datasets, we generate \textbf{6,000} synthetic images as training data.
To maintain class balance, we generate 30 training images for each class.

\noindent \textbf{CLIP backbones.} 
To demonstrate the effect of \method across various CLIP vision encoders, we conducted experiments utilizing four different CLIP backbone models\footnote{\url{https://github.com/openai/CLIP}}: 
CLIP-RN101 (ResNet~\cite{he2016deep}),
CLIP-ViT-B/16 (ViT~\cite{dosovitskiy2020image}),
CLIP-ViT-L/14,
CLIP-ViT-L/14@336px
.

\noindent \textbf{Classification.}
To demonstrate the robustness of our \method method, we leverage two different classifiers in our experiments: Logistic Regression (LR) and Multilayer perception (MLP).
For the LR classifier, we used a regularization parameter \(C=0.316\). For the Multi-Layer MLP classifier, we employed a single hidden layer with 256 neurons, ReLU activation, a learning rate of 0.001, and the Adam optimizer. Both models were configured with a maximum of 1000 iterations and a fixed random seed 42.

\vspace{-1.5mm}
\subsection{Experimental Results and Analysis}

\begin{table}[t]
\caption{Main Results for Zero-shot Domain-specific Image Classification. We evaluate \method, the base prompt, and CLIP Zero-shot across four different CLIP checkpoints. All results are reported as the average accuracy on the test set. The best result in each setting is highlighted in \textbf{bold}, with the performance improvement over the CLIP Zero-shot marked in \textcolor{black!30!green}{$\uparrow$\textbf{green}} and the performance decline marked in \textcolor{black!10!red}{$\downarrow$\textbf{red}.}}
\centering
\renewcommand{\arraystretch}{1.3}
\resizebox{\columnwidth}{!}{
\begin{tabular}{ccccc}
\hline
\rowcolor{lightgray}
\textbf{CLIP}                                             & \multicolumn{2}{c}{\textbf{Method}}                                             & \textbf{CUB-Photo}                       & \textbf{CUB-Painting}                    \\ \hline
                                                          & \multicolumn{2}{c}{CLIP Zero-Shot}                                              & 48.52\%                                  & 34.43\%                                  \\
                                                          &                                                   & LR                          & 47.85\%                                  & 29.08\%                                  \\ \cline{3-3}
                                                          & \multirow{-2}{*}{Base Prompt}                     & MLP                         & 42.51\%                                  & 30.95\%                                  \\
                                                          & \cellcolor[HTML]{E6E6E6}                          & \cellcolor[HTML]{E6E6E6}LR  & \cellcolor[HTML]{E6E6E6}\textbf{49.67\%\textcolor{black!30!green}{$\uparrow$1.15\%}} & \cellcolor[HTML]{E6E6E6}32.66\%          \\ \cline{3-3}
\multirow{-5}{*}{CLIP-RN101}                              & \multirow{-2}{*}{\cellcolor[HTML]{E6E6E6}AttrSyn} & \cellcolor[HTML]{E6E6E6}MLP & \cellcolor[HTML]{E6E6E6}44.15\%          & \cellcolor[HTML]{E6E6E6}\textbf{32.85\%\textcolor{black!10!red}{$\downarrow$1.58\%}}          \\ \hline
                                                          & \multicolumn{2}{c}{CLIP Zero-Shot}                                              & 55.11\%                                  & 39.32\%                                  \\
                                                          &                                                   & LR                          & 52.57\%                                  & 40.99\%                                  \\ \cline{3-3}
                                                          & \multirow{-2}{*}{Base Prompt}                     & MLP                         & 50.71\%                                  & 41.32\%                                  \\
                                                          & \cellcolor[HTML]{E6E6E6}                          & \cellcolor[HTML]{E6E6E6}LR  & \cellcolor[HTML]{E6E6E6}\textbf{57.97\%\textcolor{black!30!green}{$\uparrow$2.86\%}} & \cellcolor[HTML]{E6E6E6}\textbf{44.50\%\textcolor{black!30!green}{$\uparrow$5.18\%}} \\ \cline{3-3}
\multirow{-5}{*}{CLIP-ViT-B/16}                           & \multirow{-2}{*}{\cellcolor[HTML]{E6E6E6}AttrSyn} & \cellcolor[HTML]{E6E6E6}MLP & \cellcolor[HTML]{E6E6E6}54.73\%          & \cellcolor[HTML]{E6E6E6}44.31\%          \\ \hline
                                                          & \multicolumn{2}{c}{CLIP Zero-Shot}                                              & 62.65\%                                  & 42.89\%                                  \\
                                                          &                                                   & LR                          & 64.91\%                                  & 50.87\%                                  \\ \cline{3-3}
                                                          & \multirow{-2}{*}{Base Prompt}                     & MLP                         & 61.96\%                                  & 49.98\%                                  \\
                                                          & \cellcolor[HTML]{E6E6E6}                          & \cellcolor[HTML]{E6E6E6}LR  & \cellcolor[HTML]{E6E6E6}\textbf{66.86\%\textcolor{black!30!green}{$\uparrow$4.21\%}} & \cellcolor[HTML]{E6E6E6}55.46\%          \\ \cline{3-3}
\multirow{-5}{*}{CLIP-ViT-L/14}                           & \multirow{-2}{*}{\cellcolor[HTML]{E6E6E6}AttrSyn} & \cellcolor[HTML]{E6E6E6}MLP & \cellcolor[HTML]{E6E6E6}65.77\%          & \cellcolor[HTML]{E6E6E6}\textbf{56.51\%\textcolor{black!30!green}{$\uparrow$13.62\%}} \\ \hline
\multicolumn{1}{l}{}                                      & \multicolumn{2}{c}{CLIP Zero-Shot}                                              & 63.32\%                                  & 43.94\%                                  \\
\multicolumn{1}{l}{}                                      &                                                   & LR                          & 66.09\%                                  & 48.77\%                                  \\ \cline{3-3}
\multicolumn{1}{l}{}                                      & \multirow{-2}{*}{Base Prompt}                     & MLP                         & 60.80\%                                  & 50.48\%                                  \\
\multicolumn{1}{l}{}                                      & \cellcolor[HTML]{E6E6E6}                          & \cellcolor[HTML]{E6E6E6}LR  & \cellcolor[HTML]{E6E6E6}\textbf{68.73\%\textcolor{black!30!green}{$\uparrow$5.41\%}} & \cellcolor[HTML]{E6E6E6}54.12\%          \\ \cline{3-3}
\multicolumn{1}{l}{\multirow{-5}{*}{CLIP-ViT-L/14@336px}} & \multirow{-2}{*}{\cellcolor[HTML]{E6E6E6}AttrSyn} & \cellcolor[HTML]{E6E6E6}MLP & \cellcolor[HTML]{E6E6E6}66.76\%          & \cellcolor[HTML]{E6E6E6}\textbf{54.55\%\textcolor{black!30!green}{$\uparrow$10.61\%}} \\ \hline
\end{tabular}
}
\label{tab:main-results}
\end{table}

\begin{figure}[t]
  \centering
  \begin{subfigure}{0.5\linewidth}
    \includegraphics[width=\linewidth]{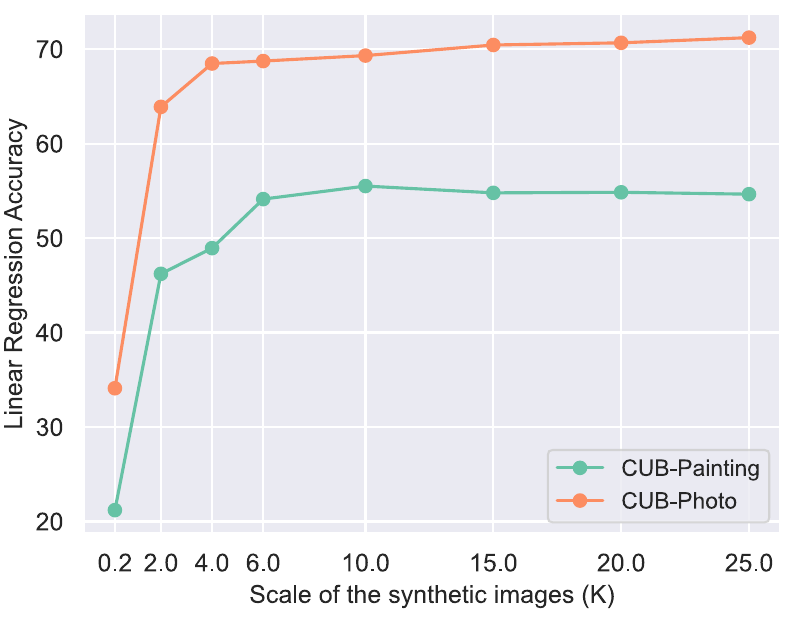}
    \caption{LR}
    \label{fig:scale-lr}
  \end{subfigure}
  \hspace{-1.5ex}
  \begin{subfigure}{0.5\linewidth}
    \includegraphics[width=\linewidth]{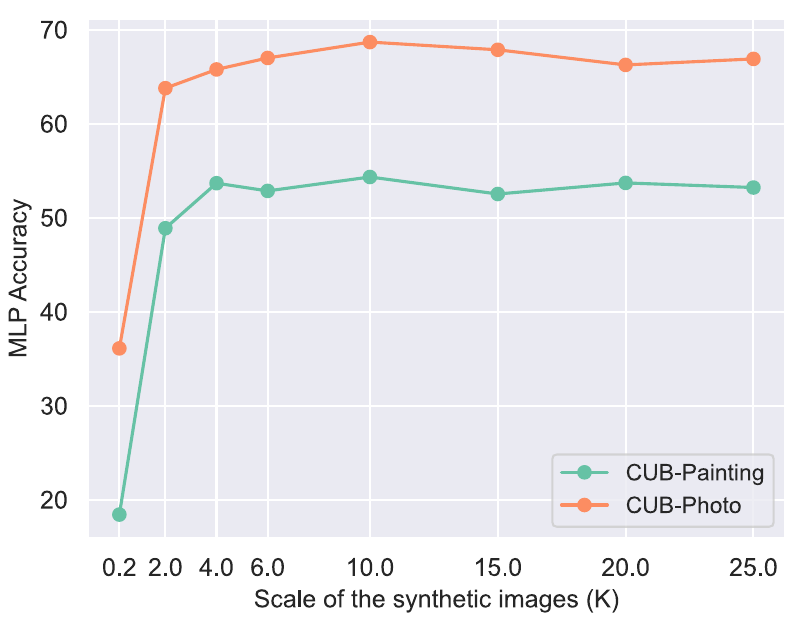}
    \caption{MLP}
    \label{fig:scale-mlp}
  \end{subfigure}
  \caption{Test performances of different scales of synthetic training data generated by our \method method.}
  \label{fig:scale}
\end{figure}

\noindent \textbf{Main results.}
The performances of models trained on synthetic images generated by the base prompt and \method are shown in Tab.~\ref{tab:main-results}, from which we can see that for both datasets, \method consistently outperforms the base prompt across all CLIP checkpoints and classifiers. Additionally, \method significantly surpasses CLIP's zero-shot performance under most situations, yielding performance enhancements ranging up to a maximum of \textbf{13.62\%}.

\noindent \textbf{Impact of different CLIP vision encoders.}
As shown in Tab.~\ref{tab:main-results}, stronger CLIP vision encoders, such as CLIP-ViT-L/14@336px, exhibit greater performance gains over CLIP zero-shot when trained on synthetic data, compared to weaker encoders. This improvement stems from their enhanced ability to generalize abstract information from synthetic images, which scales with model size. Furthermore, ViT-based CLIP vision encoders consistently outperform their ResNet-based counterparts. Synthetic images often lack realistic local features, making ResNet's reliance on localized convolutional patterns less effective, leading to performance degradation. In contrast, the Vision Transformer's global attention mechanism excels at capturing high-level, abstract features, making it particularly effective for handling synthetic data.

\noindent \textbf{Impact of different classifiers.}
As shown in Tab.~\ref{tab:main-results}, Linear Regression (LR) often outperforms Multi-Layer Perceptron (MLP) when trained on synthetic image embeddings. The simplicity of LR helps it avoid overfitting to the noise and artifacts typically present in synthetic data, whereas MLP’s higher capacity makes it more susceptible to overfitting. Since CLIP’s vision encoders already provide rich feature embeddings, LR’s straightforward approach enables better generalization, making it more effective for classification tasks using synthetic data compared to MLP in most cases.

\noindent \textbf{Ablation of synthetic data scales.}
To evaluate the impact of varying scales of synthetic training images on test performances, we generated synthetic training images based on \method in quantities of 200, 2k, 4k, 6k, 10k, 15k, 20k, and 25k for both CUB-200 (Photo) and CUB-200-Painting datasets. Subsequently, we evaluate the performances of LR and MLP using CLIP-ViT-L/14@336px as the CLIP vision encoder.
The experimental result curve is shown in Fig.~\ref{fig:scale}, from which we can see that as the scale of synthetic training images increases, the test performances improve but eventually reach a plateau. In the future, we consider it significant to investigate efficient selection strategies for synthetic training images.

\vspace{-1.5mm}
\section{Conclusion}

In conclusion, we explore synthetic image generation aiming to achieve more effective zero-shot domain-specific image classification. We introduce \method, a novel synthetic image generation method that leverages large language models to generate attributed prompts. These prompts enable the generation of attributed images with greater diversity to narrow the gap between synthetic and real images. The experiments conducted on two fine-grained datasets demonstrate the significant improvements of our \method over the base prompt and the CLIP zero-shot classification, highlighting its potential in zero-shot domain-specific image classification tasks.

\bibliographystyle{IEEEbib}
\bibliography{icme2025references}

\begin{thebibliography}{10}

\bibitem{deng2009imagenet}
Jia Deng, Wei Dong, Richard Socher, Li-Jia Li, Kai Li, and Li~Fei-Fei,
\newblock ``Imagenet: A large-scale hierarchical image database,''
\newblock in {\em 2009 IEEE conference on computer vision and pattern recognition}. Ieee, 2009, pp. 248--255.

\bibitem{mann2020language}
Ben Mann, N~Ryder, M~Subbiah, J~Kaplan, P~Dhariwal, A~Neelakantan, P~Shyam, G~Sastry, A~Askell, S~Agarwal, et~al.,
\newblock ``Language models are few-shot learners,''
\newblock {\em arXiv preprint arXiv:2005.14165}, 2020.

\bibitem{radford2019language}
Alec Radford, Jeffrey Wu, Rewon Child, David Luan, Dario Amodei, Ilya Sutskever, et~al.,
\newblock ``Language models are unsupervised multitask learners,''
\newblock {\em OpenAI blog}, vol. 1, no. 8, pp. 9, 2019.

\bibitem{schuhmann2022laion}
Christoph Schuhmann, Romain Beaumont, Richard Vencu, Cade Gordon, Ross Wightman, Mehdi Cherti, Theo Coombes, Aarush Katta, Clayton Mullis, Mitchell Wortsman, et~al.,
\newblock ``Laion-5b: An open large-scale dataset for training next generation image-text models,''
\newblock {\em Advances in Neural Information Processing Systems}, vol. 35, pp. 25278--25294, 2022.

\bibitem{wang2019survey}
Wei Wang, Vincent~W Zheng, Han Yu, and Chunyan Miao,
\newblock ``A survey of zero-shot learning: Settings, methods, and applications,''
\newblock {\em ACM Transactions on Intelligent Systems and Technology (TIST)}, vol. 10, no. 2, pp. 1--37, 2019.

\bibitem{menon2022visual}
Sachit Menon and Carl Vondrick,
\newblock ``Visual classification via description from large language models,''
\newblock {\em arXiv preprint arXiv:2210.07183}, 2022.

\bibitem{pratt2023does}
Sarah Pratt, Ian Covert, Rosanne Liu, and Ali Farhadi,
\newblock ``What does a platypus look like? generating customized prompts for zero-shot image classification,''
\newblock in {\em Proceedings of the IEEE/CVF International Conference on Computer Vision}, 2023, pp. 15691--15701.

\bibitem{yang2023diffusion}
Ling Yang, Zhilong Zhang, Yang Song, Shenda Hong, Runsheng Xu, Yue Zhao, Wentao Zhang, Bin Cui, and Ming-Hsuan Yang,
\newblock ``Diffusion models: A comprehensive survey of methods and applications,''
\newblock {\em ACM Computing Surveys}, vol. 56, no. 4, pp. 1--39, 2023.

\bibitem{nichol2021improved}
Alexander~Quinn Nichol and Prafulla Dhariwal,
\newblock ``Improved denoising diffusion probabilistic models,''
\newblock in {\em International conference on machine learning}. PMLR, 2021, pp. 8162--8171.

\bibitem{song2020score}
Yang Song, Jascha Sohl-Dickstein, Diederik~P Kingma, Abhishek Kumar, Stefano Ermon, and Ben Poole,
\newblock ``Score-based generative modeling through stochastic differential equations,''
\newblock {\em arXiv preprint arXiv:2011.13456}, 2020.

\bibitem{rombach2022high}
Robin Rombach, Andreas Blattmann, Dominik Lorenz, Patrick Esser, and Bj{\"o}rn Ommer,
\newblock ``High-resolution image synthesis with latent diffusion models,''
\newblock in {\em Proceedings of the IEEE/CVF conference on computer vision and pattern recognition}, 2022, pp. 10684--10695.

\bibitem{cai2024uncertainty}
Fiona Cai, Emily Mu, and John Guttag,
\newblock ``Uncertainty inclusive contrastive learning for leveraging synthetic images,''
\newblock in {\em Synthetic Data for Computer Vision Workshop@ CVPR 2024}, 2024.

\bibitem{li2024synthetic}
Yuhang Li, Xin Dong, Chen Chen, Jingtao Li, Yuxin Wen, Michael Spranger, and Lingjuan Lyu,
\newblock ``Is synthetic image useful for transfer learning? an investigation into data generation, volume, and utilization,''
\newblock {\em arXiv preprint arXiv:2403.19866}, 2024.

\bibitem{he2022synthetic}
Ruifei He, Shuyang Sun, Xin Yu, Chuhui Xue, Wenqing Zhang, Philip Torr, Song Bai, and Xiaojuan Qi,
\newblock ``Is synthetic data from generative models ready for image recognition?,''
\newblock {\em arXiv preprint arXiv:2210.07574}, 2022.

\bibitem{fan2023scaling}
Lijie Fan, Kaifeng Chen, Dilip Krishnan, Dina Katabi, Phillip Isola, and Yonglong Tian,
\newblock ``Scaling laws of synthetic images for model training... for now,''
\newblock {\em arXiv preprint arXiv:2312.04567}, 2023.

\bibitem{gowda2023synthetic}
Shreyank~N Gowda,
\newblock ``Synthetic sample selection for generalized zero-shot learning,''
\newblock in {\em Proceedings of the IEEE/CVF Conference on Computer Vision and Pattern Recognition}, 2023, pp. 58--67.

\bibitem{peng2015learning}
Xingchao Peng, Baochen Sun, Karim Ali, and Kate Saenko,
\newblock ``Learning deep object detectors from 3d models,''
\newblock in {\em Proceedings of the IEEE international conference on computer vision}, 2015, pp. 1278--1286.

\bibitem{chen2019learning}
Yuhua Chen, Wen Li, Xiaoran Chen, and Luc~Van Gool,
\newblock ``Learning semantic segmentation from synthetic data: A geometrically guided input-output adaptation approach,''
\newblock in {\em Proceedings of the IEEE/CVF conference on computer vision and pattern recognition}, 2019, pp. 1841--1850.

\bibitem{ros2016synthia}
German Ros, Laura Sellart, Joanna Materzynska, David Vazquez, and Antonio~M Lopez,
\newblock ``The synthia dataset: A large collection of synthetic images for semantic segmentation of urban scenes,''
\newblock in {\em Proceedings of the IEEE conference on computer vision and pattern recognition}, 2016, pp. 3234--3243.

\bibitem{abu2018augmented}
Hassan Abu~Alhaija, Siva~Karthik Mustikovela, Lars Mescheder, Andreas Geiger, and Carsten Rother,
\newblock ``Augmented reality meets computer vision: Efficient data generation for urban driving scenes,''
\newblock {\em International Journal of Computer Vision}, vol. 126, pp. 961--972, 2018.

\bibitem{yen2022nerf}
Lin Yen-Chen, Pete Florence, Jonathan~T Barron, Tsung-Yi Lin, Alberto Rodriguez, and Phillip Isola,
\newblock ``Nerf-supervision: Learning dense object descriptors from neural radiance fields,''
\newblock in {\em 2022 international conference on robotics and automation (ICRA)}. IEEE, 2022, pp. 6496--6503.

\bibitem{shipard2023diversity}
Jordan Shipard, Arnold Wiliem, Kien~Nguyen Thanh, Wei Xiang, and Clinton Fookes,
\newblock ``Diversity is definitely needed: Improving model-agnostic zero-shot classification via stable diffusion,''
\newblock in {\em Proceedings of the IEEE/CVF Conference on Computer Vision and Pattern Recognition}, 2023, pp. 769--778.

\bibitem{yu2023diversify}
Zhuoran Yu, Chenchen Zhu, Sean Culatana, Raghuraman Krishnamoorthi, Fanyi Xiao, and Yong~Jae Lee,
\newblock ``Diversify, don't fine-tune: Scaling up visual recognition training with synthetic images,''
\newblock {\em arXiv preprint arXiv:2312.02253}, 2023.

\bibitem{yang2017diversity}
Zichen Yang, Haifeng Liu, and Deng Cai,
\newblock ``On the diversity of realistic image synthesis,''
\newblock {\em arXiv preprint arXiv:1712.07329}, 2017.

\bibitem{yu2024large}
Yue Yu, Yuchen Zhuang, Jieyu Zhang, Yu~Meng, Alexander~J Ratner, Ranjay Krishna, Jiaming Shen, and Chao Zhang,
\newblock ``Large language model as attributed training data generator: A tale of diversity and bias,''
\newblock {\em Advances in Neural Information Processing Systems}, vol. 36, 2024.

\bibitem{birds2011}
Catherine Wah, Steve Branson, Peter Welinder, Pietro Perona, and Serge Belongie,
\newblock ``The caltech-ucsd birds200-2011 dataset,'' \url{http://www.vision.caltech.edu/visipedia/CUB-200-2011.html}, 2011.

\bibitem{wang2020progressive}
Sinan Wang, Xinyang Chen, Yunbo Wang, Mingsheng Long, and Jianmin Wang,
\newblock ``Progressive adversarial networks for fine-grained domain adaptation,''
\newblock in {\em Proceedings of the IEEE/CVF conference on computer vision and pattern recognition}, 2020, pp. 9213--9222.

\bibitem{radford2021learning}
Alec Radford, Jong~Wook Kim, Chris Hallacy, Aditya Ramesh, Gabriel Goh, Sandhini Agarwal, Girish Sastry, Amanda Askell, Pamela Mishkin, Jack Clark, et~al.,
\newblock ``Learning transferable visual models from natural language supervision,''
\newblock in {\em International conference on machine learning}. PMLR, 2021, pp. 8748--8763.

\bibitem{rohrbach2011evaluating}
Marcus Rohrbach, Michael Stark, and Bernt Schiele,
\newblock ``Evaluating knowledge transfer and zero-shot learning in a large-scale setting,''
\newblock in {\em CVPR 2011}. IEEE, 2011, pp. 1641--1648.

\bibitem{lampert2013attribute}
Christoph~H Lampert, Hannes Nickisch, and Stefan Harmeling,
\newblock ``Attribute-based classification for zero-shot visual object categorization,''
\newblock {\em IEEE transactions on pattern analysis and machine intelligence}, vol. 36, no. 3, pp. 453--465, 2013.

\bibitem{zhang2015zero}
Ziming Zhang and Venkatesh Saligrama,
\newblock ``Zero-shot learning via semantic similarity embedding,''
\newblock in {\em Proceedings of the IEEE international conference on computer vision}, 2015, pp. 4166--4174.

\bibitem{bauer2024comprehensive}
Andr{\'e} Bauer, Simon Trapp, Michael Stenger, Robert Leppich, Samuel Kounev, Mark Leznik, Kyle Chard, and Ian Foster,
\newblock ``Comprehensive exploration of synthetic data generation: A survey,''
\newblock {\em arXiv preprint arXiv:2401.02524}, 2024.

\bibitem{ho2020denoising}
Jonathan Ho, Ajay Jain, and Pieter Abbeel,
\newblock ``Denoising diffusion probabilistic models,''
\newblock {\em Advances in neural information processing systems}, vol. 33, pp. 6840--6851, 2020.

\bibitem{ramesh2021zero}
Aditya Ramesh, Mikhail Pavlov, Gabriel Goh, Scott Gray, Chelsea Voss, Alec Radford, Mark Chen, and Ilya Sutskever,
\newblock ``Zero-shot text-to-image generation,''
\newblock in {\em International conference on machine learning}. Pmlr, 2021, pp. 8821--8831.

\bibitem{achiam2023gpt}
Josh Achiam, Steven Adler, Sandhini Agarwal, Lama Ahmad, Ilge Akkaya, Florencia~Leoni Aleman, Diogo Almeida, Janko Altenschmidt, Sam Altman, Shyamal Anadkat, et~al.,
\newblock ``Gpt-4 technical report,''
\newblock {\em arXiv preprint arXiv:2303.08774}, 2023.

\bibitem{podell2023sdxl}
Dustin Podell, Zion English, Kyle Lacey, Andreas Blattmann, Tim Dockhorn, Jonas M{\"u}ller, Joe Penna, and Robin Rombach,
\newblock ``Sdxl: Improving latent diffusion models for high-resolution image synthesis,''
\newblock {\em arXiv preprint arXiv:2307.01952}, 2023.

\bibitem{he2022masked}
Kaiming He, Xinlei Chen, Saining Xie, Yanghao Li, Piotr Doll{\'a}r, and Ross Girshick,
\newblock ``Masked autoencoders are scalable vision learners,''
\newblock in {\em Proceedings of the IEEE/CVF conference on computer vision and pattern recognition}, 2022, pp. 16000--16009.

\bibitem{he2016deep}
Kaiming He, Xiangyu Zhang, Shaoqing Ren, and Jian Sun,
\newblock ``Deep residual learning for image recognition,''
\newblock in {\em Proceedings of the IEEE conference on computer vision and pattern recognition}, 2016, pp. 770--778.

\bibitem{dosovitskiy2020image}
Alexey Dosovitskiy,
\newblock ``An image is worth 16x16 words: Transformers for image recognition at scale,''
\newblock {\em arXiv preprint arXiv:2010.11929}, 2020.

\end{thebibliography}

\end{document}